%% file: main.tex
\documentclass[conference]{IEEEtran}
\usepackage[T1]{fontenc}
\usepackage{newtxtext}

\usepackage{graphicx}
\usepackage{booktabs}
\usepackage{array}
\usepackage{amsmath}
\usepackage{newtxmath}
\usepackage{url}
\usepackage[hidelinks]{hyperref}
\usepackage{cite}
\hypersetup{pdftitle={A Case Study of a Multi-Agent Kernel-Optimization Loop That Learns a New AI Accelerator},pdfauthor={Byonggon Chun}}

\begin{document}
\bstctlcite{IEEEexample:BSTcontrol}

\title{A Case Study of a Multi-Agent Kernel-Optimization Loop That Learns a New AI Accelerator}

\author{\IEEEauthorblockN{Byonggon Chun}
\IEEEauthorblockA{Fluidstack\\
byonggon@fluidstack.io}}

\maketitle

\begin{abstract}
\input{sections/00_abstract}
\end{abstract}

\input{sections/01_intro}
\input{sections/02_background}
\input{sections/08a_compare_table}
\input{sections/02b_setup}
\input{sections/03_onboarding}
\input{sections/04_loop}

\input{sections/05_results}

\input{sections/06_failed}
\input{sections/07_noise}
\input{sections/09_conclusion}

\bibliographystyle{IEEEtran}
\bibliography{refs}

\end{document}

%% file: sections/00_abstract.tex
Frontier large language models write correct CUDA kernels for most single-operator benchmark tasks, yet their accuracy collapses on accelerators whose architecture and programming language they have rarely seen. We present a case study of an LLM agent loop built for such accelerators. Our example is the FuriosaAI RNGD NPU, programmed in a recent, sparsely documented Rust DSL, where the loop optimized three Gemma-4-12B decode kernels over 12 days on a shared, scarce device. The loop builds its own target knowledge: it writes and checks a summary of the DSL documentation, measures the chip with microbenchmarks, records compiler-derived legality rules, and revisits this knowledge when the toolchain changes, rechecking rules and measurements where it can. A role-based search over per-kernel Design Trees explores whole-kernel designs and accumulates small accepted changes in the best version of each kernel. Because device time is scarce and readings are noisy, cheap checks screen new designs before any device run, and late in the run a change was accepted only after repeated comparisons with the best kernel on the device. The final kernels reach a 9.293$\times$ geomean speedup over the official baseline, each within 1.25$\times$ of the cycles needed just to move its weights. Our prompts, profiler and acceptance test are available at \url{https://github.com/bgchun-fs/multi-agent-kernel-gen-loop}.

%% file: sections/01_intro.tex
\section{Introduction}\label{sec:intro}

Frontier large language models (LLMs) have become capable kernel programmers on mature platforms. On CUDA, the best of them produce functionally correct kernels for most KernelBench Level 1 tasks, single operators such as matrix multiplication, within a few tries \cite{kernelbench,evokernel}, and agentic systems built on them tune kernels for accelerators such as Gemmini and AWS Trainium \cite{autocomp,accelopt}. This strength does not carry over to data-scarce targets: when these tasks move to Ascend C, an NPU language with little public code, GPT-5.2's few-shot correctness (pass@4) drops from 92\% on CUDA to 14\% \cite{evokernel}. A new domain-specific accelerator can start from exactly this cold start, with a new memory hierarchy, a new programming language and almost no expert kernels to learn from.

We study this cold start on one such target, in a live setting. In Round 1 of the NPU Model Optimization Competition at MICRO 2026 \cite{moasite}, we spent 12 days speeding up three decoder kernels of Gemma-4-12B \cite{gemma} on the FuriosaAI RNGD, a Tensor Contraction Processor \cite{tcp} programmed in furiosa-opt, a recent Rust DSL \cite{furiosabook}. RNGD is a fitting example: its DSL is sparsely documented, and no curated knowledge of the chip or the DSL existed for prompts. The device was shared and scarce, reached only through FuriosaAI's Arena job scheduler, and the toolchain and the grader changed during the round. Existing LLM kernel optimizers start from human-written target knowledge in their prompts \cite{autocomp,accelopt} or learn across many tasks \cite{evokernel}; a new target has no such knowledge, and speeding up one model may give only a few tasks. JAXBench points to what must come first: on the sparsely documented Pallas DSL, curated documentation raised per-sample correctness from 5.8\% to 37.3\% with no change to the search \cite{jaxbench}.

Each part of our loop answers one constraint that such a target imposes. With no prompt-ready knowledge of the target, \emph{onboarding} (Figure \ref{fig:onboard}) builds and checks a target knowledge base: a verified summary of the DSL book, device facts measured by microbenchmarks, and compiler-derived legality rules, rechecked when the toolchain changes. With few kernels to optimize, the loop searches deeply within each: per-kernel \emph{Design Trees} hold whole-kernel designs and their variants, and the \emph{head}, the best accepted version of each kernel, accumulates the levers (kinds of change) that pass the acceptance test, one or several at a time. With a shared, scarce device, the loop spends device time sparingly: new designs pass cheap checks before their first device run, races stop losing variants early, and the acceptance test can stop after a few pairs. With noisy readings, a profiler ties each change to the device timeline, and an \emph{acceptance test} compares a candidate with the head over repeated device runs. Roles that keep notebooks and budgets (Figure \ref{fig:loop}) carry what the loop learns across short-lived agents. In our case, the round gave three kernels and two device jobs per account (Section \ref{sec:setup}).

The final kernels reach speedups of 3.75$\times$, 11.17$\times$ and 19.16$\times$ over the official baseline cycles, a 9.293 geomean that placed 7th of 79 teams on the public leaderboard (Section \ref{sec:results}). Each takes at most 1.25$\times$ its peak, the fewest cycles needed just to move its weights (Section \ref{sec:setup-peak}). Our contributions are:
\begin{itemize}
\item \textbf{Onboarding as an explicit phase} that builds and rechecks target knowledge for a new chip and DSL (Section \ref{sec:onboard}).
\item \textbf{A role-based loop over per-kernel Design Trees} for an unfamiliar accelerator, with its checks ordered around a scarce device (Section \ref{sec:loop}).
\item \textbf{Device-grounded judging under noise,} with what our acceptance test could not do (Sections \ref{sec:loop} and \ref{sec:noise}).
\item \textbf{A record of what failed} in a live competition (Section \ref{sec:fail}).
\end{itemize}
Two approaches failed: the compiler's static schedule did not predict small gains on the device, and the loop proposed test-input shortcuts, which we accepted at first, then banned, and exclude from all results (Section \ref{sec:fail}).

%% file: sections/02_background.tex
\section{Background and Related Work}\label{sec:related}

\subsection{The Example Target: RNGD}\label{sec:setting}

Our example target is the FuriosaAI RNGD NPU, a Tensor Contraction Processor \cite{tcp} with two clusters that a kernel can use together; the host launches the two clusters one after the other, so the second starts later, by a delay we call the launch skew. Kernel weights stream from the chip's high-bandwidth memory (HBM). RNGD is programmed in furiosa-opt, a Rust DSL in which the programmer manages memory placement and Tensor Unit scheduling directly, while the furiosa-opt compiler inserts the synchronization and orders the work by data dependencies \cite{furiosabook}. The chip and the DSL are recent and sparsely documented, so we assume that our models had seen little of either.

\subsection{Related Work}

\paragraph{LLM kernel agents} KernelBench \cite{kernelbench} made kernel writing a benchmark task, and AlphaEvolve \cite{alphaevolve} evolves programs, including a kernel tiling heuristic, through LLM edits. Early in the run, our loop took a block of target rules, a repair turn after each compile error and a partial plan-then-implement step from Autocomp \cite{autocomp}, the contrast between slow and fast kernels from AccelOpt \cite{accelopt}, and anti-cheating checks before compile from EvoKernel \cite{evokernel}; its role split resembles MetaGPT \cite{metagpt} and ChatDev \cite{chatdev}. Table \ref{tab:compare} compares our loop with the three closest kernel agents.

\paragraph{Autotuning} The Design Tree draws mainly on autotuning. Ansor \cite{ansor} separates a program's structure (a sketch) from its numbers (annotations such as tile sizes), and MetaSchedule \cite{metaschedule} composes search spaces from transformation modules; our variant generators follow Ansor's split, and the loop races variants on the device, following irace \cite{irace} and Hyperband \cite{hyperband}. Unlike these systems, it searches designs and generators written by LLM agents.

\paragraph{Measurement noise} Mytkowicz et al. \cite{mytkowicz} show that harmless-looking setup details can bias measured speed, and duet benchmarking \cite{duet} runs two versions at once so that shared noise hits both. Our acceptance test borrowed this idea by submitting the candidate and the head in pairs, but on Arena the two readings of a pair proved uncorrelated (Section \ref{sec:noise}). AutoTuneBench \cite{autotunebench} and a real-workload study \cite{realworkload} build trustworthy measurement for tuning loops.

\paragraph{Verification and cheating} Kernel benchmarks invite reward hacking. The AI CUDA Engineer \cite{aicuda} found a memory exploit that let some kernels skip correctness checks, robust-kbench \cite{robustkbench} and KernelBench-Verified \cite{kbverified} found further loopholes, including bypasses hard-coded for specific test values, and KernelGuard \cite{kernelguard} protects live GPU kernel competitions. We add a case from the participant side: shortcuts built on a competition's fixed test inputs, which our loop proposed and we accepted at first, then banned (Section \ref{sec:fail-cheat}).

%% file: sections/08a_compare_table.tex
\begin{table*}[t]
\centering
\caption{Design comparison of our loop with the three closest LLM kernel agents, from their papers. Because the targets and tasks differ, the table compares designs only.}\label{tab:compare}
\footnotesize
\begin{tabular}{>{\raggedright\arraybackslash}p{1.8cm}>{\raggedright\arraybackslash}p{3.45cm}>{\raggedright\arraybackslash}p{3.45cm}>{\raggedright\arraybackslash}p{3.45cm}>{\raggedright\arraybackslash}p{3.45cm}}
\toprule
Aspect & AccelOpt \cite{accelopt} & Autocomp \cite{autocomp} & EvoKernel \cite{evokernel} & Ours \\
\midrule
Target & AWS Trainium (NKI), also H100 (Triton) & Gemmini, AWS Trainium, NVIDIA L40S & NPU (Ascend C), also CUDA & FuriosaAI RNGD NPU (furiosa-opt Rust DSL), shared and scarce \\
Target knowledge & NKI basics and a programming guide in the prompt; lessons from slow-fast pairs & ISA in the prompt (dropped for the GPU), a menu of optimizations & Data-scarce NPU language; memory seeded with API templates & Built by agents (DSL summary, microbenchmarks, legality rules); rechecked after a toolchain change \\
Unit of search & A candidate kernel in the beam & One optimization per step: plan, then implement & A first draft, then refinements & Whole-kernel designs and their variants; one head per kernel \\
Search & Beam search: keep 6, 144 new candidates per round, 16 rounds & Beam search (6), menu dropout, LLM ensembling & Memory-based reinforcement learning with value-driven retrieval & Design Tree; cheap checks before device runs; variants raced \\
Measurement & Latency on the chip via profiling; \% of peak & Real hardware, or a cycle-accurate simulator for Gemmini & Anti-hacking, compile and correctness gates; profiler latency & Per-cluster device traces; repeated acceptance test against the head \\
\bottomrule
\end{tabular}
\end{table*}

%% file: sections/02b_setup.tex
\section{Experimental Setup}\label{sec:setup}

\subsection{Task and Score}\label{sec:setup-task}

We evaluate our agent loop in Round 1 of the NPU Model Optimization Competition, held with the MOA Workshop at MICRO 2026 \cite{moasite}. The round asked teams to speed up three decode kernels of one Gemma-4-12B decoder layer \cite{gemma}, each processing one token: the QKV projection, the attention output projection and the feed-forward block, which we call \texttt{qkv}, \texttt{attn\_out} and \texttt{ffn}. One call streams 31.5, 15.7 and about 99.5 MB of weights, respectively, and weights and inputs are synthetic random values. The score is the geometric mean, over the three kernels, of the official baseline cycles (Table \ref{tab:results}) divided by our cycles.

\subsection{Graders and the Toolchain Change}\label{sec:setup-graders}

The final grader, which we call the seeded grader, runs each kernel on three input cases from fixed, public seeds, which differ in activations, weights, scales and RoPE position \cite{moarepo}; it grades the median of the three runs, each timed on the slower cluster without host time, and checks every output against the organizers' PyTorch reference. It arrived mid-round with a new toolchain release, furiosa-opt 0.8.1, which the organizers recommended over 0.6.0; we moved our kernels to 0.8.1 and repeated onboarding (Section \ref{sec:onboard}). The old grader timed one run on one fixed input case across both clusters. The official baselines date from the earlier toolchain and grader, where they match our run of the unmodified baseline within 1.2\%; on the final grader, the organizers' unmodified 0.8.1 baseline scores 0.9623. The public board shows each team's best graded row for reference only; the official evaluation re-runs each team's last submission.

\subsection{Device Access}\label{sec:setup-device}

Teams reach the device only through Arena, FuriosaAI's job scheduler for shared RNGD servers, which allows each account at most two active jobs of at most 70 s. The loop and its compiles ran on an EC2 build host.

\subsection{Agents and Run}\label{sec:setup-agents}

Claude agents ran claude-opus-5-5 and GPT agents ran gpt-6-astra, both at their highest reasoning effort. Our run lasted 12 days (283 h), from 09-16 to 09-28, 2026.

\subsection{Peak}\label{sec:setup-peak}

A kernel's peak is the fewest cycles needed just to move its weights on this device: 54.6k, 29.0k and 174.0k cycles for \texttt{qkv}, \texttt{attn\_out} and \texttt{ffn}, measured with probe kernels that only move the weights on both clusters (for \texttt{qkv}, estimated from its 31.5 MB at the measured two-cluster rate plus fixed costs). The peak is a planning reference, not a proven bound.

%% file: sections/03_onboarding.tex
\section{Onboarding the Agent}\label{sec:onboard}

\begin{figure}[t]
\centering
\includegraphics[width=0.94\columnwidth]{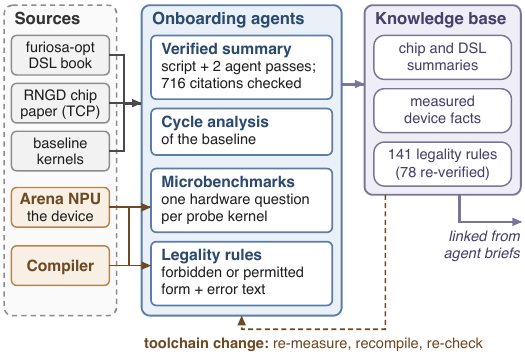}
\caption{Onboarding. Agents turn the DSL book, the chip paper, the baseline kernels, the device and the compiler into a knowledge base of verified summaries, measured device facts and 141 legality rules. Agent briefs link to it, and the phase is repeated when the toolchain changes.}
\label{fig:onboard}
\end{figure}

We assume that an agent cannot search well on a target it cannot yet program correctly, since a search can rank only kernels that compile and compute correctly. Onboarding therefore runs before the search and builds the knowledge base of Figure \ref{fig:onboard}.

\paragraph{Verified summary} Agents wrote a cited summary of the furiosa-opt book \cite{furiosabook}, the DSL's manual, and checked it against the book with a script and two agent passes: of its 716 citations, 683 fully supported their claims, none contradicted the book, and the agents fixed the other 33. They also wrote a cited summary of the TCP paper \cite{tcp}, a cycle analysis of the baseline kernels and a four-part kernel explanation whose parts each passed a GPT evaluation.

\paragraph{Microbenchmarks} Documents describe the chip, but only the device shows what it does. A microbenchmark is a small standalone kernel that answers one hardware question on the device, such as how fast a weight load from HBM really runs. One probe that only moves a 15.7 MB weight tensor from HBM ran only 1.47 times as fast on both clusters as on one, so a resource that the two clusters share limits the rate. This shared rate, not twice the one-cluster rate, sets the peak that the strategist targets (Section \ref{sec:setup-peak}), and the final kernels split weight streams between the clusters and limit exchanges between them (Section \ref{sec:results}).

\paragraph{Legality rules} Compile failures and device findings became legality rules in a DSL cheat sheet. Each rule names a forbidden or permitted form, and a rule for a compile refusal quotes its exact error text, such as \texttt{VRU reduce axes must be innermost in Partitioning}, so an agent can find the rule from the error and an architect can check a design before writing it.

\paragraph{Onboarding again} When the organizers changed the toolchain and the grader mid-round (Section \ref{sec:setup-graders}), we paused the loop, moved our kernels to the new release and repeated onboarding: a sweep tool planted each forbidden form in our kernels and recompiled them, a rule whose error text still matched became verified, and device rules were re-measured. In the end, the cheat sheet held 141 legality rules: 78 re-verified, 16 rewritten, 2 retired and 45 carried over unverified, because nothing on the new toolchain confirmed or contradicted them.

%% file: sections/04_loop.tex
\section{Loop Architecture}\label{sec:loop}

Our loop searches both whole-kernel designs and small changes to the best kernel, judges both on the device, and keeps knowledge in roles that outlast short-lived agents.\footnote{The prompts of every role, the profiler and the acceptance test are at \url{https://github.com/bgchun-fs/multi-agent-kernel-gen-loop}.}

\begin{figure*}[t]
\centering
\includegraphics[width=0.95\textwidth]{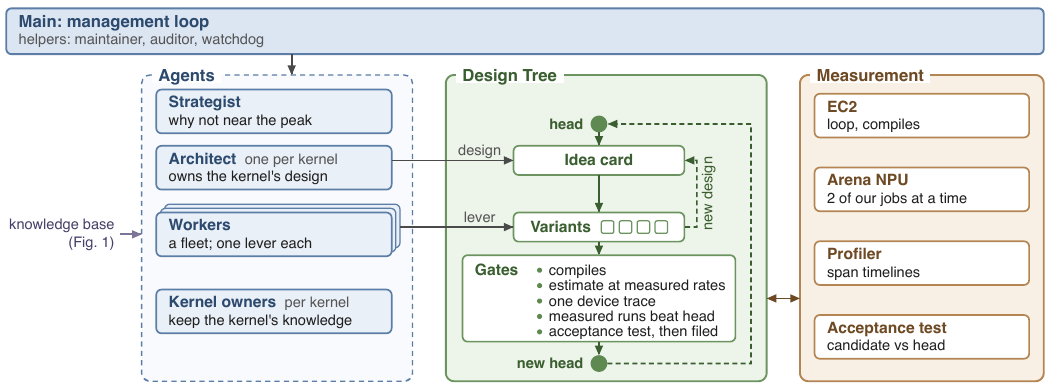}
\caption{The agent loop. Main, the management loop, runs the agents (left), which grow each kernel's Design Trees (middle): architects add whole-kernel designs as idea cards, and workers add variants, including one-lever changes to the head, our current best accepted kernels. A measured variant can also seed a new design (dashed), so a lineage mixes the two. Directly written variants pass the gates in order, generated variants are raced, and a one-lever change goes to the acceptance test; a filed winner becomes the new head. The loop and its builds run on EC2; device runs, profiler timelines and acceptance tests use the Arena NPU (right).}
\label{fig:loop}
\end{figure*}

\subsection{Roles}\label{sec:loop-roles}

Figure \ref{fig:loop} shows the roles.
\begin{itemize}
\item \textbf{Main}, the management loop, is the only role that spawns and stops agents, applies decisions and files submissions to the grader. Its helpers are a maintainer agent for tool changes, auditor agents, and a watchdog script that checks the loop every 30 minutes.
\item \textbf{Strategist:} a special run of main every 3 h or when the watchdog flags a stall. It asks why the kernels are not yet near their peak (Section \ref{sec:setup-peak}) and whether the approach is wrong, ranks levers by expected gain and by evidence, stops workers on dead levers and writes new briefs; it builds nothing.
\item \textbf{Architect:} owns one kernel's design, writes a budget of where its cycles go and proposes structurally different designs as idea cards for up to 4 workers to build. When the kernel runs out of ideas, main starts a new architect iteration that redesigns it from the hardware budget instead of the current code.
\item \textbf{Worker:} one agent with one brief, which names a kernel and one lever (one kind of change) with its evidence and target. It builds, times on the device and reports; levers that pass the acceptance test join the head, one or several at a time, so accepted levers accumulate. Up to 36 agents run at once.
\item \textbf{Kernel owner:} keeps one kernel's notebook, a running record of that kernel's knowledge that outlasts each worker, and reviews the workers' changes.
\end{itemize}
The loop is designed to operate autonomously. Agents wrote all of the loop's code and all of the kernels, at the operator's request or with the operator's approval. The human operator can intervene by submitting a task file to the loop; for example, near the end of the run, the operator asked the loop to stop exploring and to refine its best candidates for submission.

\subsection{Design Trees}\label{sec:loop-tree}

Each kernel grows a forest of Design Trees, 22 to 34 of them, each rooted in one idea card. A node is a whole-kernel design, a distinct structure for one kernel, or a variant, a concrete kernel that is built and measured. A new design starts as an architect's idea card that states the design and a kill criterion, the trace result that would end the idea; a variant carries one or more accepted levers stacked on its design, and a new design often branches from a measured variant, so a lineage mixes the two and can run deep: in \texttt{ffn} one runs eight nodes deep. Each kernel keeps one head across its trees instead of a beam of alternatives. Early in the run, the loop used beam search and accepted a change only with a gain of at least 3\%, first in the static schedule (Section \ref{sec:fail}) and then on the device; most single changes gained less (on \texttt{ffn}, 1.5-3\%), so few passed. This motivated accepting smaller, device-validated changes on a single head, where accepted levers accumulate. Variants come from three sources, each with its own path to the head, and each path spares the scarce device in its own way:
\begin{itemize}
\item \textbf{Written directly} by workers from an idea card, then checked by six gates in order: (1) the card defines the design and its arithmetic; (2) the kernel compiles within 3 tries; (3) before any device time, an estimate at measured device rates predicts at least 3\% over the head; (4) one device trace confirms the predicted change; (5) a batch of device runs beats the head; (6) main promotes the winner.
\item \textbf{Generated} by a variant generator, a small program that an agent writes and a script runs, which renders one kernel per combination of its settings; following Ansor \cite{ansor}, each setting declares structure (a ``sketch'') or a tunable choice (an ``annotation'', such as a tile order). Generated variants are raced: each is read on the device against repeated readings of a reference kernel, thresholds in standard deviations of those readings pick the winners, and a winner must still pass the acceptance test.
\item \textbf{One lever:} a worker changes one lever of the head and takes it straight to the acceptance test, which may stop after 5 pairs.
\end{itemize}
Searching whole designs lets the loop leave a design whose small changes have stopped paying off. Generated variants added little: of 337 rendered, 4 won their races and one of them passed the acceptance test.

\subsection{Device-Grounded Judging}\label{sec:loop-judge}

Our profiler joins a kernel's device timeline, the timed spans that the harness records on each cluster, with the compiler's static schedule, so that an agent can aim a change at one slow span and check it afterwards; as in Perfetto's trace processor \cite{perfetto}, agents query the traces with SQL. The launch skew changes from run to run, so on the old grader, whose window spanned both clusters, the acceptance rule used program time with the skew removed and later compared traces only at matched skew.

Late in the run, on the seeded grader, the loop accepts a candidate through the acceptance test, a repeated comparison with the head. The test submits the candidate and the head in pairs, one device job of each, one second apart and in alternating order, and each job reads every kernel as the grader does. It computes $g = \ln(\text{candidate score} / \text{head score})$ per pair, and the candidate passes when $z$, the mean of $g$ over 11 pairs divided by its standard error, is at least 1.64; a test may stop after 5 pairs when $z$ is above 4 or below $-2$. A filing also needs an estimated gain of at least 0.5\%.

%% file: sections/05_results.tex
\section{Results}\label{sec:results}

Table \ref{tab:results} reports a geomean speedup of 9.293, with each final kernel at 1.11-1.25$\times$ its peak (Section \ref{sec:setup-peak}), so the speedups differ mainly because the baselines sat at different distances from their peaks: 4.6$\times$ for \texttt{qkv}, 14.0$\times$ for \texttt{attn\_out} and 21.3$\times$ for \texttt{ffn}.

\begin{table}[t]
\centering
\caption{Final graded submission. Final cycles are the seeded grader's median of 3 runs; baseline cycles are the official values from the earlier grader. Peak is the planning reference of Section \ref{sec:setup-peak}.}
\label{tab:results}
\small
\setlength{\tabcolsep}{4pt}
\begin{tabular}{lrrrrr}
\toprule
Kernel & Baseline & Final & Speedup & Peak & Final / \\
 & cycles & cycles & ($\times$) & cycles & peak \\
\midrule
qkv & 250,514 & 66,817 & 3.75 & 54,600 & 1.22 \\
attn\_out & 404,633 & 36,227 & 11.17 & 29,000 & 1.25 \\
ffn & 3,703,473 & 193,246 & 19.16 & 174,000 & 1.11 \\
\midrule
Geomean & & & 9.293 & & \\
\bottomrule
\end{tabular}
\end{table}

\begin{figure}[t]
\centering
\includegraphics[width=0.94\columnwidth]{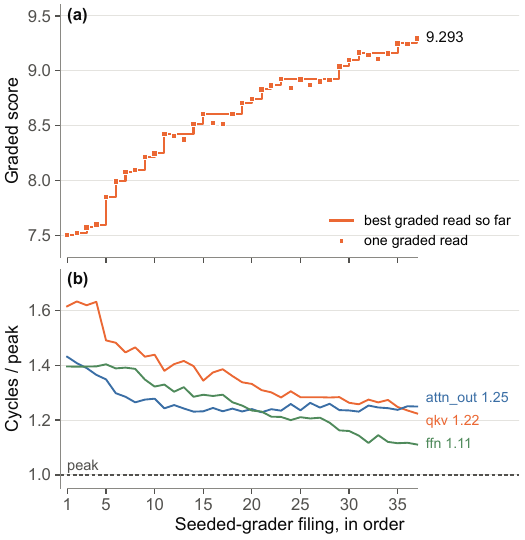}
\caption{Our filings on the updated toolchain and the seeded grader, in order (Sep 22 to 28, 2026), without the 3 rows that used the test-input shortcuts of Section \ref{sec:fail-cheat} and one control filing of the unmodified baseline. (a) Graded score: the line is the best graded read so far, and dots below it are reads that did not beat it, including re-reads of identical trees. (b) Each kernel's device cycles divided by its peak (Section \ref{sec:setup-peak}); 1.0 is the peak.}
\label{fig:score}
\end{figure}

\paragraph{Where the gain came from} Figure \ref{fig:score} traces the gain on the seeded grader alone, since old-grader scores are on a different scale; two graded reads of one tree differed by up to 1.0\%. From the clean head of 7.52 (clean rows carry no test-input shortcut; Section \ref{sec:fail-cheat}), the largest single steps in acceptance tests were a \texttt{qkv} redesign ($-7{,}907$ cycles; filing 5 in Figure \ref{fig:score}b), a split of the \texttt{ffn} down projection ($-6{,}900$) and gathering both clusters' \texttt{attn\_out} row sums onto one cluster ($-1{,}510$).

\paragraph{A worked example} An audit of the \texttt{ffn} head found three round trips to HBM and a down projection cut into small striped commands. A worker's brief named one lever: each cluster multiplies its own half of the intermediate activation by its half of the down-projection weights, so the activation needs no exchange between clusters and the two partial outputs are summed once at the end. The first build moved only 390-450 bytes per cycle against 554 on the head; rebuilt to move whole half-rows of the weights in large commands, it moved 519-533. The acceptance test passed it over 11 pairs ($z = 8.07$), and the seeded grader moved \texttt{ffn} from 241,250 to 234,468 cycles.

\paragraph{Ranking} The organizers published no official ranking. Our 9.293 row was also our last submission, the one that the official evaluation re-runs. On the public board, it ranked 5th of 75 when our run ended and, after two teams passed it before the deadline, 7th of 79 on 10-04 and 10-14; the teams ranked 6th and 8th scored within 0.4\% of it, inside the grader's spread. Our team advanced to Round 2.

%% file: sections/06_failed.tex
\section{What Did Not Work}\label{sec:fail}

Two approaches failed during the run, and each shaped a part of the loop described in Section \ref{sec:loop}.

\subsection{Trusting the Compiler's Static Schedule}

Early in the run, the loop ranked candidates by the compiler's static makespan, the cycle count of its planned schedule: by default, a candidate reached the device only with a static gain of at least 3\%, and one or two device runs decided acceptance. The planned schedule proved a poor guide to gains: four \texttt{qkv} candidates with a static gain of 6.7-14\% gained at most 2.7\% on the device, and the 18 accepted changes took 1.75-2.38$\times$ their planned time on the device (median 2.0), a ratio that drifted as the kernels changed. Each real time is a single device run and, for changes on both clusters, includes the launch skew (Section \ref{sec:setting}), which the plan does not model. The static schedule tracked large structural changes but missed the small gains the loop was chasing, so in the loop, device runs rank and accept candidates, and the static schedule only filters them.

\subsection{Folds: A Shortcut We Count as Cheating}\label{sec:fail-cheat}

A fold hard-codes a value or property of the competition's fixed test inputs, such as the RoPE position (137), the fact that one activation was exactly $\pm 1$, scale literals, or statistics computed offline from the inputs. A folded kernel skips work that the reference does and stays correct only while the inputs keep that value or property. The loop proposed folds, and we accepted them to see how far they could raise our score under the grading at the time.

The seeded grader's new inputs exposed most folds: by our reading of the organizers' new harness and input generator, 6 of the 10 fold switches we had built would fail. We banned folds, allowed the 4 surviving ones once more for a short window in which three fold rows were graded, and then banned folds for good by an explicit rule: nothing in a kernel may stand in for a quantity of the test inputs. The organizers' change notes describe this class of shortcut \cite{moarepo}, and their switch to the seeded grader closed most of it for all teams. We did not report our fold rows to the organizers; they stayed on the board until higher fold-free rows replaced them. All 44 fold rows are left out of our results.

Our checks before compile did not catch folds: their planned rule for constants was never built, and the model judge that stood in for it was told to let folds through once we had accepted them. Since the ban, every brief carries the rule and the filing script rejects the known fold switches and long new float literals. We call a row clean when it contains no fold and no value taken from the test inputs. The final kernels are clean: instead of hard-coding the RoPE position, as one fold did, they compute the RoPE tables on the chip from the position with the reference's formula, which matches the reference within one bf16 rounding step at every position from 0 to 32,767. Clean rows up to a score of 9.04 carried \texttt{ffn} scaling constants sized from the old grader's input case and the real checkpoint; these set numeric headroom and replace no computed value.

We count folds as cheating: a folded kernel does not compute its function. A loop under score pressure can propose such shortcuts, so the rule against them belongs in its first brief, together with test inputs that vary from the start.

%% file: sections/07_noise.tex
\section{Measurement Noise}\label{sec:noise}

On a shared and scarce device, any single reading is one draw. Arena did not report which device ran a job, and we had no dedicated device to compare with, so we could not tell whether the noise came from the Arena environment or from the RNGD chip and its software. The environmental factors we could test on the seeded grader showed no measurable effect: our other job running at the same time, other teams' load, the hour and the queue wait each moved readings by at most 70 cycles.

\paragraph{From the chip and runtime} We expect the main source to be the runtime's launch path, through the launch skew (Section \ref{sec:setting}). On the old grader, whose window spanned both clusters, the skew added a median of 12.5k cycles to \texttt{attn\_out} (6.4k to 17.8k from the 10th to the 90th percentile of 384 traces), and no program change moved its distribution; the seeded grader, which reads each cluster's own window, sees it only through waits between clusters, about 1-2k cycles on \texttt{attn\_out}. On \texttt{qkv}, the first graded run, which always uses the same input case, was the slowest in all 37 filings checked, so run order and input case could not be separated.

\paragraph{How large it was} On the old grader, re-runs of one binary minutes apart differed, as $(\max-\min)/\min$, by a median of 0.5\% on \texttt{qkv} and 0.8\% on \texttt{ffn} but 8.5\% on \texttt{attn\_out} (up to 12.6\%). On the seeded grader, the head's readings within an acceptance test varied with a median coefficient of variation of 0.70\% on \texttt{qkv}, 0.64\% on \texttt{attn\_out} and 0.24\% on \texttt{ffn}.

\paragraph{What the acceptance test cannot do} First, submitting the candidate and the head in pairs did not cancel shared noise: their deviations within a pair were uncorrelated (correlation $-0.001$ over 1,354 pairs), and the two jobs ran at the same time in only 35\% of the pairs we could time, so the test worked as a repeated comparison in balanced order, not as a duet benchmark. Second, a single test of 5 to 11 pairs passes a candidate with no real gain about 7-9\% of the time under a $t$ distribution, not 5\%, and early stopping and repeated tests add to this; the 0.5\% minimum gain guards against such passes. Third, the test had no ``slower'' outcome in practice, so slower candidates got the same ``flat'' verdict as noise, though some were truly slower: of the 110 flat verdicts in 174 tests, 19 had $z \le -1.64$, against about 9 to 14 expected by chance, and four had $z < -8$.

%% file: sections/09_conclusion.tex
\section{Limitations and Conclusion}\label{sec:limits}

\paragraph{Limitations}
This study covers one chip, one competition and one run, with three decode kernels and two model families. We ran no ablations, so we cannot say which part of the loop produced the gain, and our loop never ran without its knowledge base, so we cannot say how much onboarding added. We could not tell whether the measurement noise came from the shared Arena servers or from the chip and its software (Section \ref{sec:noise}), and our acceptance test passes a candidate with no real gain more often than its nominal 5\%. The peak is a planning reference, not a proven bound, and the public-board rank is not an official ranking. Finally, we tested the final kernels only on the organizers' generated inputs, not with the real model's weights.

\paragraph{Conclusion}
We presented an LLM agent loop that first learns an unfamiliar NPU and its DSL and then optimizes its kernels in a live competition, where it reached a 9.293 geomean speedup. Our run shows that an agent loop can write and optimize kernels for a recent, sparsely documented accelerator and its DSL, with a knowledge base that it builds itself and with device runs deciding which candidates it accepts. We draw four lessons, each a hypothesis from one run. First, a loop facing a new accelerator should build and check its own knowledge of the chip and the DSL before it searches, and recheck it when the toolchain changes. Second, until device runs calibrate it, a compiler's static schedule can filter candidates but cannot judge them. Third, the acceptance rule is a component to design and to check against data, with a ``slower'' outcome and small-sample statistics. Finally, a stable, isolated measurement environment and a grader that draws varied inputs matter: noise on our shared device called for a repeated statistical acceptance test, and fixed test inputs invited the shortcuts that our loop proposed.